\documentclass[letterpaper, 10 pt, conference]{ieeeconf}  

\IEEEoverridecommandlockouts                              

\usepackage{amsmath} 
\usepackage{amssymb}  
\usepackage{units}
\usepackage{graphicx}
\usepackage{subcaption}	
\usepackage{adjustbox}
\usepackage[font=footnotesize, format=plain, labelfont=bf]{caption}
\usepackage{hyperref}

\begin{document}

\title{\LARGE \bf
Multi-Objective Global Path Planning for Lunar Exploration \\ With a Quadruped Robot
}

\author{Julia Richter, Hendrik Kolvenbach, Giorgio Valsecchi, and Marco Hutter
\thanks{This work was supported by the European Space Agency (ESA) 
under contract no. 4000135310/21/NL/PA/pt and the European Union's Horizon 2020 research and innovation programme under grant agreement No 101016970.}
\thanks{All the authors are with the Robotic Systems Lab (RSL) at ETH Zurich, Switzerland {\tt\small jurichter@ethz.ch}}%
\thanks{$^1$ \url{https://github.com/leggedrobotics/lunar_planner}} %
}

\maketitle
\thispagestyle{empty}
\pagestyle{empty}

\begin{abstract}

In unstructured environments the best path is not always the shortest, but needs to consider various objectives like energy efficiency, risk of failure or scientific outcome.
This paper proposes a global planner, based on the A* algorithm, capable of individually considering multiple layers of map data for different cost objectives.
We introduce weights between the objectives, which can be adapted to achieve a variety of optimal paths.
In order to find the best of these paths, a tool for statistical path analysis is presented.
Our planner was tested on exemplary lunar topographies to propose two trajectories for exploring the Aristarchus Plateau. The optimized paths significantly reduce the risk of failure while yielding more scientific value compared to a manually planned paths in the same area.
The planner and analysis tool are made open-source$^1$ in order to simplify mission planning for planetary scientists.

\begin{keywords}
Planetary exploration, Mission planning, Intelligent and autonomous space robotics systems
\end{keywords}

\end{abstract}

\section{INTRODUCTION}
Mankind has a history of exploring celestial bodies close to Earth - from the first unmanned mission to explore the Moon in 1959 to the latest Mars 2020 mission. Most autonomous platforms used for these missions were wheeled rovers. Their many advantages include effortless, stable balance and easy-to-control locomotion. 
However, they have challenges overcoming high slopes, rock fields, and sand dunes \cite{ono2018mars}. This prevents them from accessing some of the most interesting areas on Mars and the Moon, like impact craters or fields of volcanic ejecta.
Over the last years, legged robots gained increasing interest on Earth. Their ability to traverse a broad range of terrains \cite{wellhausen2021rough}, makes them a good candidate for future extra-terrestrial exploration.

When traversing unstructured terrains, the best path is not always the shortest path. Instead, longer paths might be justified by, e.g., a lower energy consumption or a lower risk of failure. 
While this already complies with wheeled robots, their reduced locomotion skills limit the complexity of this problem. 
In contrast, for legged robots finding an optimal path that complies with different objectives, poses a difficult optimization task that can hardly be solved manually.

\section{RELATED WORK}
\begin{figure}
    \centering
    \begin{subfigure}[b]{.65\linewidth}
        \centering
        \includegraphics[width=\linewidth]{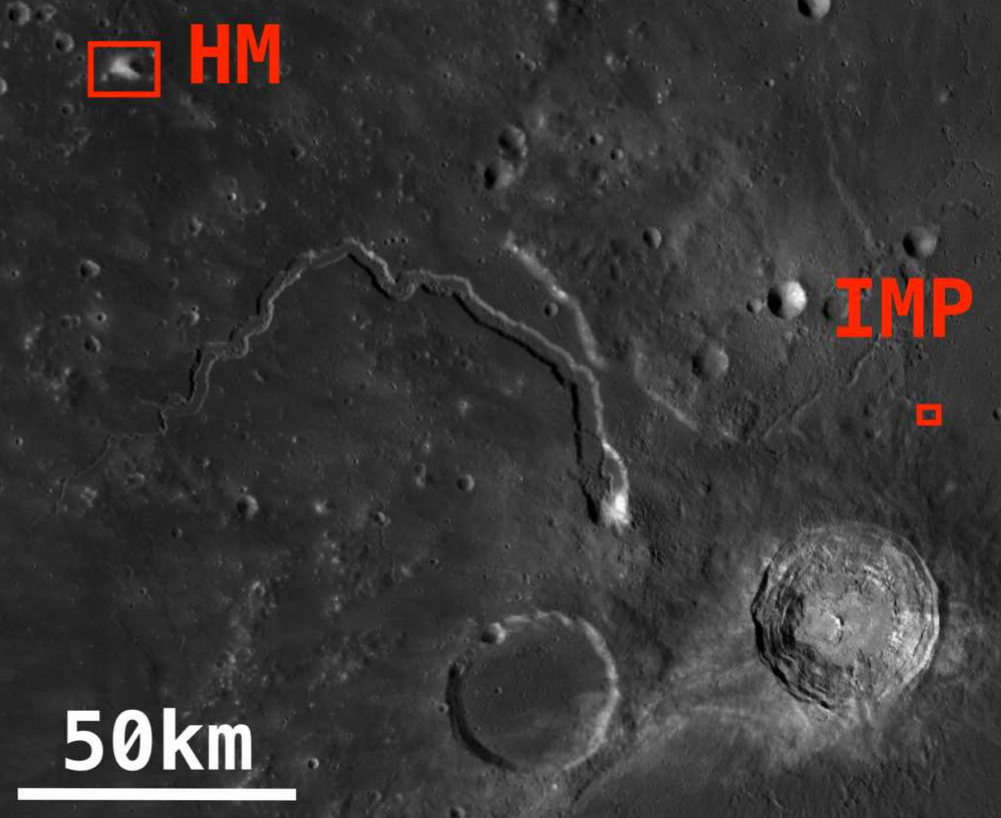}
        \caption{Overview}
        \label{fig:aristarchusoverview}
    \end{subfigure}

    \begin{subfigure}[b]{.48\linewidth}
        \centering
        \includegraphics[width=.88\linewidth]{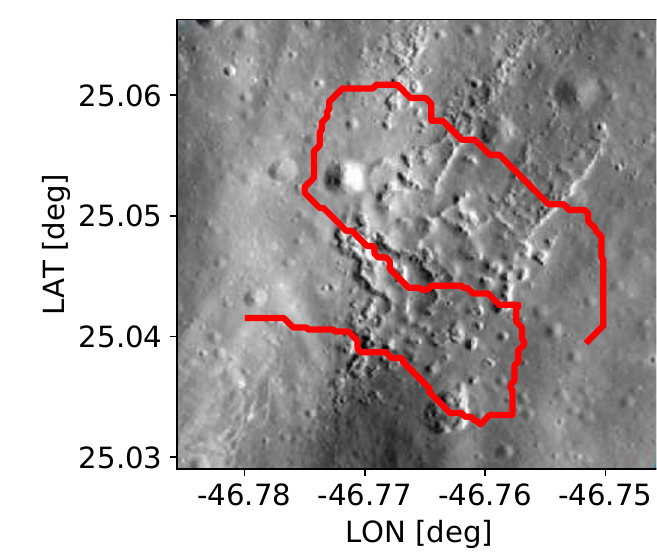}
        \caption{IMP: Manual}
        \label{fig:imp_previous}
    \end{subfigure}
    \hfill
    \begin{subfigure}[b]{.48\linewidth}
        \centering
        \includegraphics[width=\linewidth]{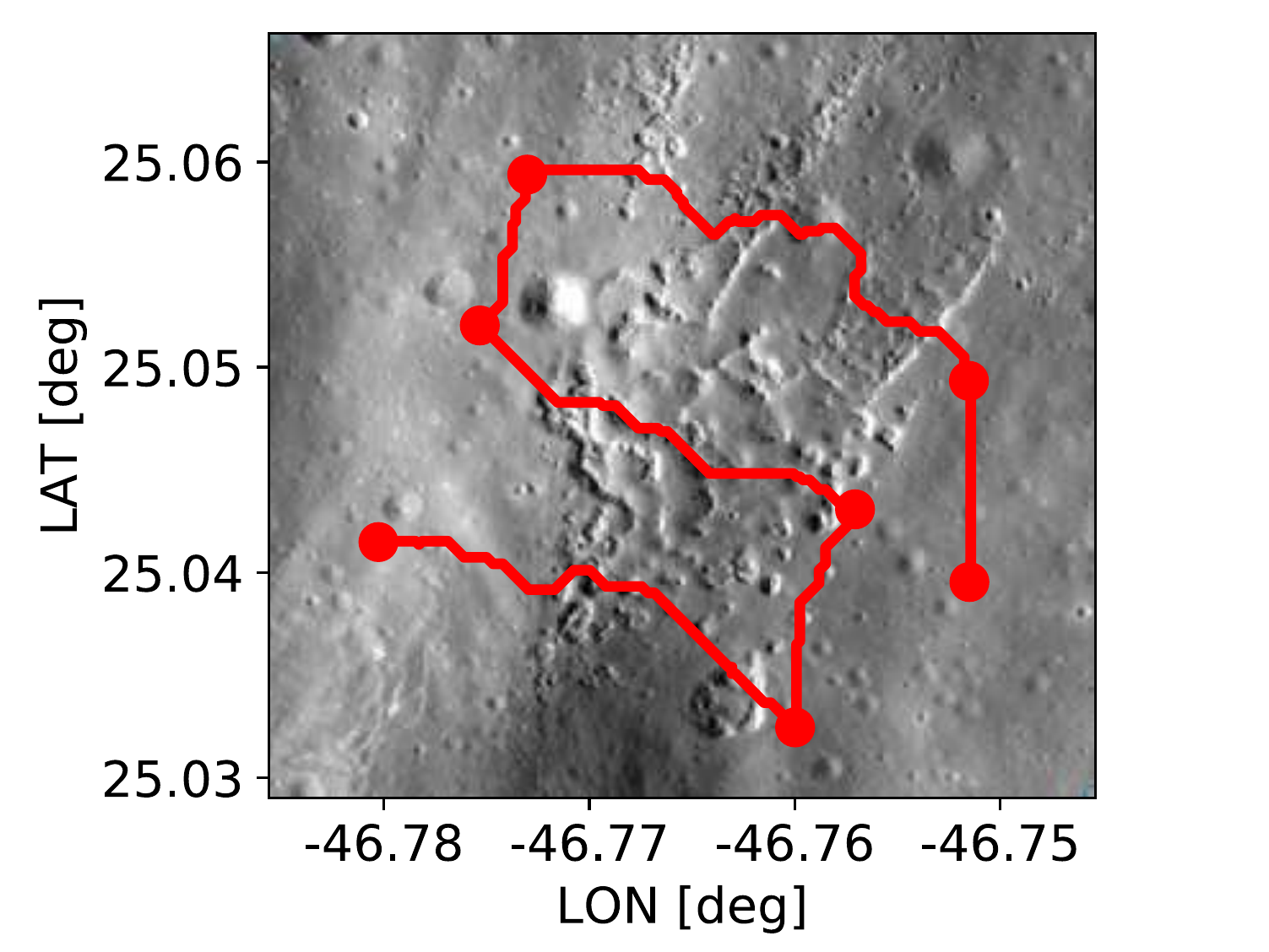}
        \caption{IMP: Tool-assisted}
        \label{fig:imp_proposed}
    \end{subfigure}

    \begin{subfigure}[b]{.48\linewidth}
        \centering
        \includegraphics[width=.98\linewidth]{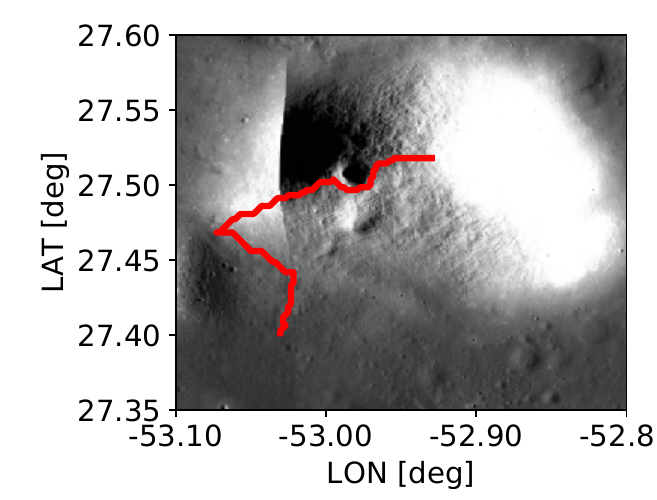}
        \caption{HM: Manual}
        \label{fig:hm_previou}
    \end{subfigure}
    \hfill
    \begin{subfigure}[b]{.48\linewidth}
        \centering
        \includegraphics[width=\linewidth]{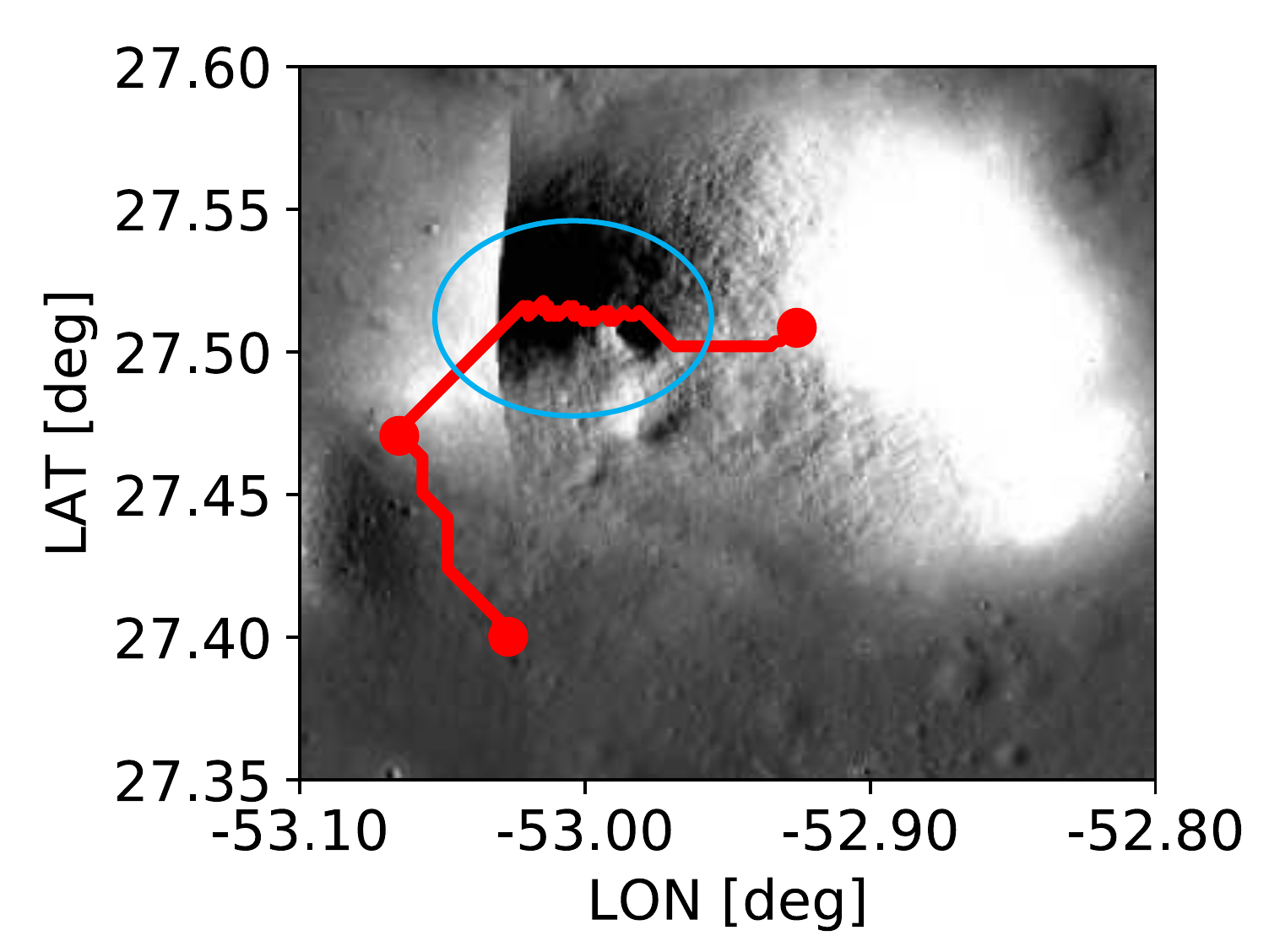}
        \caption{HM: Tool-assisted}
        \label{fig:hm_proposed}
    \end{subfigure}

    \caption{Satellite images of scientifically interesting regions around the Aristarchus crater, featuring Aristarchus IMP (\textbf{IMP}) and Herodotus Mons (\textbf{HM}).
    Figures b and d show paths created by a lunar geology expert, while the paths in Figures c and e were generated using our proposed path planning tool.}
    \label{fig:all_applications_sat_pics}
\end{figure}

There has been extensive research on path planning algorithms, which is comprehensively discussed by Sánchez-Ibáñez et al. \cite{sanchez2021path}. However, most works focus either on finding any feasible path or optimize over the shortest distance from a start to a goal position. 
The challenge of optimizing a path on a global km-scale over several objectives due to the properties of unstructured terrain has been mostly discussed in the area of extra-terrestrial exploration.

Ono et al. \cite{ono2016data} categorize the speed of Mars rovers into four categories, ranging from \unit[0]{m/Sol} to \unit[200]{m/Sol}. The term \emph{Sol} hereby refers to one day on Mars, which is the equivalent of \unit[24.6]{h} on Earth \cite{marsday}. Based on this, several papers discuss the fusion of information about the slope and ground composition to create a map of predicted speeds over which the global path can be optimized \cite{hedrick2020terrain, azkarate2020gnc, sanchez2019dynamic}. However, this approach has the drawback of the cost not being dependent on the direction the robot traverses across a terrain patch.
While this simplification may be acceptable for rovers that primarily operate on flat terrain, legged robots are capable of navigating steeper slopes as demonstrated by Kolvenbach et al. \cite{kolvenbachtowards} and Weibel et al. \cite{weibel2023towards}. In such cases, the cost relies on the direction of slope traversal.

Other approaches to include several objectives have been made by Candela and Wettergreen \cite{candela2022approach}, who introduced the idea of constraining the risk the rover takes while optimizing the scientific value.
Rao et al. \cite{rao2023multi} take this one step further and use risk and energy consumption as additional objectives instead of constraints. In their method, one optimal path is calculated using the Pareto optimality, meaning no other feasible solution can improve one objective without degrading at least one other objective. This approach was also tested in other applications than planetary exploration (\cite{bradley1991multiobjective, martins2022improved, gul2021meta}). Wells et al. \cite{wells2022optimal} introduce the idea of enabling the user to weigh in between two objectives, namely energy consumption and risk.

This work builds onto that idea by implementing a global planner targeting the challenge of finding an optimal path in unstructured environments. Global is here defined by a path length of several hundred meters up to several kilometers. Our main contributions are:
\begin{itemize}
    \item A multi-objective global planning pipeline that fuses map information and robot capabilities
    \item A tool that provides optimal paths and statistical evaluation based on user input
    \item An open-source framework allowing extension and usage for future mission planning
\end{itemize}

We use the planner to propose paths for the exploration of two representative sites on the Moon. The missions are located at Herodotus Mons (\textbf{HM}) and the Aristarchus Irregular Mare Patch (\textbf{IMP}), which are highlighted in Figure \ref{fig:aristarchusoverview}.
Both show various volcanic features on challenging terrain, according to Glotch et al. \cite{glotch2021scientific}. This justifies the exploration with legged robots instead of traditional rovers. 
The scientific relevance of Herodotus Mons lies in the in-situ analysis of diverse volcanic materials. The primary objective for the Aristarchus IMP is to investigate boulders, which are visible in the satellite image.

\section{METHODS}
Our method is based on three prerequisites. Firstly, we utilize the increasingly accurate satellite data readily accessible for the Moon. Secondly, by analysing this data with application knowledge, landing sites and scientific targets can be identified. Lastly, to plan specific mission paths, the assets and constraints of the robot need to be known and included in the path-planning process.

We assume a hierarchical path planning pipeline comprising two modules. The global planner, the focus of this work, operates on a kilometer-scale map with \unit[5-60]{m} resolution, producing a coarse path of waypoints. The local planner functions within a \unit[5-50]{m} range at \unit[5-10]{cm} resolution, using local perception information to navigate between waypoints.

Based on the utilization of satellite data, the configuration space for the global planner can only include the robot's x and y positions, as well as the yaw rotation. Assuming that the local planner handles the local pose of the robot, the yaw orientation on a global level will be neglected. This results in a two-dimensional configuration space.

The requirements for the new planner are to enable a \emph{multi-objective} as well as a \emph{directional} cost function.
Using multiple objectives to plan one final path originates from the necessity to consider diverse targets, like minimizing energy or risk or maximizing scientific outcome or illumination times. The directionality of the cost function is needed for projecting the correlation between locomotion cost and the direction of slope traversal.
The planner should further output an \emph{optimal path} and show \emph{deterministic} behavior to ensure repeatability during the mission planning process. 

Based on these requirements and the overview of Sánchez-Ibáñez et al. \cite{sanchez2021path}, we decided to use the A* graph search algorithm \cite{hart1968formal} with a custom cost $g(x)$ and heuristic function $h(x)$. 
This algorithm coincides with the requirements of being optimal and deterministic. 
However, it is memory- and run-time-intensive for larger configuration spaces and maps. While the primer is confined to two dimensions and, therefore, should not present issues, the severity of the latter will be evaluated at a later point in this paper.

To build the graph, the grid size is derived from the satellite data. 
Each resulting node is positioned at the center of a pixel and connected to its eight neighbors.
The cost function $g(\hat{x}_n)$ calculates the total cost of reaching the current node $\hat{x}_n=(x_n,y_n)$ from the start node by summing the costs of each node visited along the path. The heuristic function $h(\hat{x}_n)$ estimates the remaining cost to the goal, guiding the search direction. During each step of the A* search algorithm, the node that minimizes the function
\begin{equation}
    f(\hat{x}_n) = g(\hat{x}_n) + h(\hat{x}_n)
    \label{eq:f_gplush}
\end{equation}
is selected for further exploration.

\subsection{Customized Cost Function}
We define the total cost of the current node $\hat{x}_n$ as
\begin{equation}
    g(\hat{x}_n) = \sum_{i=1}^{n} g(\hat{x}_i)
\end{equation}
with the partial cost $g(\hat{x}_i)$ of each node $i$ along the path. 

\begin{figure}
    \centering
    \includegraphics[width=\linewidth]{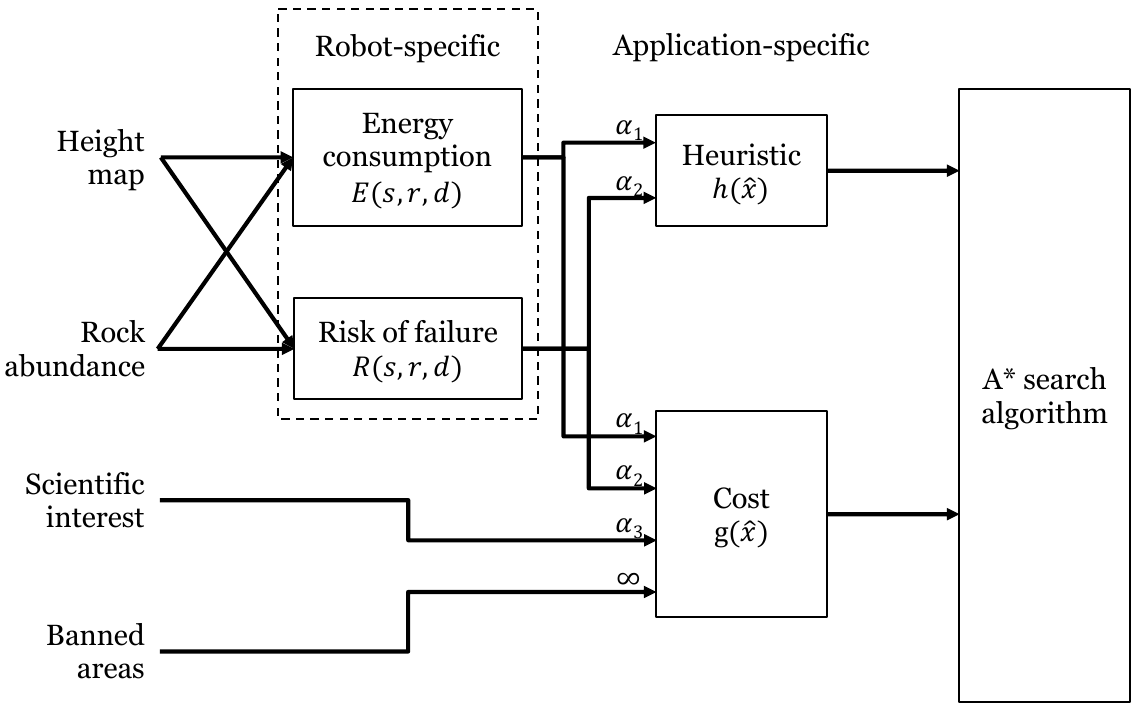}
    \caption{Global path calculation pipeline with multi-objective cost}
    \label{fig:multiobjectivecalc}
\end{figure}

For the further course of this paper, one specific use case will be discussed. The planning pipeline is presented in Figure \ref{fig:multiobjectivecalc}. The following objectives are considered:
\begin{itemize}
    \item Minimize energy consumption
    \item Minimize risk of failure
    \item Maximize scientific value
\end{itemize}
We define the risk of failure as the statistically estimated likelihood that collisions with boulders or steep slopes will cause the robot to crash, as described by Wells et al. \cite{wells2022optimal}.

The calculation is based on four distinct layers of multi-modal map information. This includes a Digital Elevation Model (\textbf{DEM}), which provides detailed terrain elevations. The rock abundance measures the size and density of rocks on the Moon, as introduced by Bandfield et al. \cite{bandfield2011lunar}. Additionally, scientific interest is quantified on a scale from 0 to 1, indicating the desirability of visiting specific locations. Finally, banned areas define where the robot is allowed or not allowed to go.

Based on these objectives, we conclude the following application-specific cost function for each node $i$:
\begin{align}
    g(\hat{x}_i, \hat{x}_{i-1}) &= \alpha_1 \cdot E(\hat{x_i}, \hat{x}_{i-1}) + \alpha_2 \cdot R(\hat{x_i}, \hat{x}_{i-1}) \nonumber \\
    &+ \alpha_3 \cdot I(\hat{x}_i) + B(\hat{x}_i)
    \label{eq:gx}
\end{align}
with the robot-specific costs for energy consumption $E(\hat{x_i}, \hat{x}_{i-1})$, risk of failure $R(\hat{x_i}, \hat{x}_{i-1})$, scientific value $I(\hat{x}_i)$ and a penalty for crossing banned areas $B(\hat{x}_i)$.
While the latter two are only dependent on the current node, the energy consumption and risk of failure are dependent on the traversed slope, hence they rely on the height values from both the current $\hat{x}_i$ and the previous node $\hat{x}_{i-1}$. The user-defined weights $\alpha_1$, $\alpha_2$ and $\alpha_3$ determine the emphasis of each objective. They must lie within $0 \leq \alpha_m \leq 1$ and accord to $\sum \alpha_m = 1, m \in \{1,2,3\}$.

\subsection{Specific Cost Components}
The modelling of the cost components depends on the used robotic platform.
We focus our experiments on the quadruped robot ANYmal \cite{hutter2016anymal}. 
In the following, the variables $E, R, B, I$ are defined in the cost space, while $E^*, R^*, B^*, I^*$ denote the corresponding variables in the physical space.

We estimate the cost for energy efficiency based on the assumption that the actuators' winding losses account for a significant component of the total energy consumption. The winding losses scale quadratically to the torque. In conclusion, achieving the minimum squared torque corresponds approximately to the lowest energy consumption.

Based on this assumption, Wells et al. \cite{wells2022optimal} model the squared torque on different slopes and boulder fields. Specifically, they run a simulation of thousands of ANYmal robots in lunar gravity on different slopes and boulder fields, with a fixed velocity of \unit[0.8]{m/s} and a grid size of $d_{sim} = \unit[8]{m}$.
The squared torque $E^*(s_i,r_i,d_i)$ while traversing an edge between the previous node $\hat{x}_{i-1}$ and the current node $\hat{x}_{i}$ over a distance $d_i=d(\hat{x_i}, \hat{x}_{i-1})$ dependent on a rock abundance $r_i=r(\hat{x}_i)$ and a slope $s_i=s(\hat{x_i}, \hat{x}_{i-1})$ is:
\begin{equation}
    E^*(s_i,r_i,d_i) = (p_0 + p_1 s_i + p_2 r_i + p_3 s_i^2 + p_4 s_i r_i + p_5 r_i^2) \frac{d_i}{d_{sim}}
    \label{eq:wells1}
\end{equation}
The polynomial coefficients $p_j$ are defined in Table \ref{tab:coefficients}.
To ensure comparability of the various cost components from Equation \ref{eq:gx}, we scale Equation \ref{eq:wells1} to the maximal possible value $E^*_{max}$ based on the map-specific combination of maximal slope and rock abundance.
This leads to the cost component $E(\hat{x_i}, \hat{x}_{i-1})$ for node $\hat{x_i}$:
\begin{equation}
    E(\hat{x_i}, \hat{x}_{i-1}) = \frac{E^*(s_i,r_i,d_i)}{E^*_{max}}
    \label{eq:e}
\end{equation}
with $E(\hat{x_i}, \hat{x}_{i-1}) \in [E^*_{min}/E^*_{max}, 1]$.

\begin{table}[]
    \centering
    \begin{tabular}{r|cc}
        $p_j$ & Energy $E(\hat{x_i}, \hat{x}_{i-1})$ & Crash rate $R(s_i,r_i)$ \\
        \hline
        $p_0$ & $803$ & $-2.88e^{-2}$ \\
        \hline
        $p_1$ & $10.5$ & $5.31e^{-4}$ \\
        \hline
        $p_2$ & $70.3$ & $0.319$ \\
        \hline
        $p_3$ & $0.739$ & $3.14e^{-4}$ \\
        \hline
        $p_4$ & $-1.42$ & $-2.3e^{-2}$ \\
        \hline
        $p_5$ & $1770$ & $10.8$ \\
    \end{tabular}        
    \caption{Coefficients for squared torque and crash rate as modelled by Wells et al. \cite{wells2022optimal}}
    \label{tab:coefficients}
\end{table}

The cost for the risk $R(x_i, x_{i-1})$ is derived from the crash rate $R^*(s_i,r_i)$, which is modelled by Wells et al. \cite{wells2022optimal} as
\begin{equation}
    R^*(s_i,r_i) = p_0 + p_1 s_i + p_2 r_i + p_3 s_i^2 + p_4 s_i r_i + p_5 r_i^2
    \label{eq:wells3}
\end{equation}
with the polynomial coefficients $p_j$ as defined in Table \ref{tab:coefficients}.
Since this estimate accounts only for a traversed distance of $d_{sim}=\unit[8]{m}$, it needs to be scaled to the actual distance $d_i$ with
\begin{equation}
    R^*_{scaled}(s_i,r_i,d_i) = 1-(1-R^*(s_i,r_i))^{d_i/d_{sim}}
    \label{eq:r_scaled}
\end{equation}
As before, the cost component $R(x_i, x_{i-1})$ is calculated with
\begin{equation}
    R(x_i, x_{i-1}) = \frac{R^*_{scaled}(s_i,r_i,d_i)}{R^*_{max}}
    \label{eq:r}
\end{equation}
The scientific value of one node is defined as
\begin{equation}
    I(\hat{x}_i) = 1 - I^*(\hat{x}_i)
    \label{eq:costinterest}
\end{equation}
where $I^*(\hat{x}_i)$ is the normalized value from the map \emph{Scientific interest}. This formula assigns a low cost to nodes with a high scientific interest. While introducing a negative cost for the scientific value might be more intuitive, since scientific discoveries justify, e.g., a higher risk, a negative cost would also compromise the ability of A* to find the optimal path.

The penalty for crossing banned areas is realized through
\begin{equation}
    B(\hat{x}_i) =  \begin{cases}
                        \infty, & B^*(\hat{x}_i)=1\\
                        0, & B^*(\hat{x}_i)=0
                    \end{cases}    
\end{equation}
where $B^*(\hat{x}_i)$ is the value from the map \emph{Banned areas}. To prevent the planner from exploring regions that are beyond the robot's locomotion capabilities, areas with a slope or a rock abundance exceeding the maxima of $\pm\unit[30]{deg}$ and $0.3$, respectively, are added to the banned areas. 

\subsection{Heuristic Function}

The heuristic function $h(\hat{x}_n)$ guides the search to the goal by introducing an estimate of the remaining cost to reach the goal. Being part of the Equation \ref{eq:f_gplush}, it directs the search towards nodes that are closer to the goal.

To ensure that the path with the least cost is generated, the heuristic function needs to have two characteristics: It should be \emph{admissible}, which means to not overestimate the cost for reaching the goal, and \emph{consistent}. A heuristic with these characteristics leads according to Hart et al. \cite{hart1968formal} to fewer nodes being visited and hence a faster path calculation.

The applied heuristic function includes the costs for energy efficiency and risk of failure, since only these parts of the cost are distance dependent.
The heuristic function $h(\hat{x}_n)$ is defined as
\begin{equation}
    h(\hat{x}_n) = h_{min} \cdot d(\hat{x}_n, \hat{x}_{goal})
\end{equation}
with the map-specific minimal cost of traversing one edge:
\begin{equation}
    h_{min} = \alpha_1 \cdot E_{min} + \alpha_2 \cdot R_{min}
\end{equation}
While $\alpha_1$ and $\alpha_2$ are the weights from Equation \ref{eq:gx}, $E_{min}$ and $R_{min}$ describe the map-specific minimal possible costs for energy efficiency and risk of failure.
To demonstrate the \emph{admissibility} and \emph{consistency} of this heuristic, a proof following the approach of Wells et al. \cite{wells2022optimal} can be done.

\subsection{Path Evaluation}

While evaluating paths based on the final cost components provides one measure, transforming these cost components back to the physical world enhances interpretability.
For the energy efficiency, this can be done with the scaling factor from Equation \ref{eq:e}.
The approximated relative energy $E^*_{total}$ required to traverse $n$ nodes, with respective costs $E_i=E(\hat{x_i}, \hat{x}_{i-1})$, is
\begin{equation}
    E^*_{total} = \sum_{i=1}^{n} E_i \cdot E^*_{max} \cdot t
    \label{eq:etotal}
\end{equation}
Here, $t$ represents the time needed to traverse each node and takes values from $\{10, \sqrt{2}\times 10\}$ s, depending on whether the traversal is straight or diagonal.
The absolute crash risk $R^*_i$ for one node $i$ with cost $R_i=R(\hat{x_i}, \hat{x}_{i-1})$ based on Equations \ref{eq:r_scaled} and \ref{eq:r} is
\begin{equation}
    R^*_i = 1 - \sqrt[d/d_{sim}]{1-R_i \cdot R^*_{max}}
\end{equation}
The total crash risk $R^*_{total}$ is
\begin{equation}
    R^*_{total} = 1 - \prod_{i=1}^{n} \Bigl( 1-R^*_i \Bigr)
    \label{eq:rtotal}
\end{equation}
This function is derived from the likelihood of traversing the full path without the robot encountering a crash.

The scientific outcome can be estimated by analysing the percentage of path elements that are of scientific interest. 
Since the scientific value $I_{total}^*$ is not dependent on the distance covered, but on the visited grid fields, it is scaled to the number of nodes $n$ that are visited. It is calculated with 
\begin{equation}
    I_{total}^* = 1 - \frac{\sum_{i=1}^{n} I_{i}}{n}
    \label{eq:itotal}
\end{equation}
where $I_i=I(\hat{x_i})$ is the scientific cost of Equation \ref{eq:costinterest}.

\subsection{Statistical Analysis} \label{sec:methods_analysis}
\begin{figure}
    \centering
    \includegraphics[width=.6\linewidth]{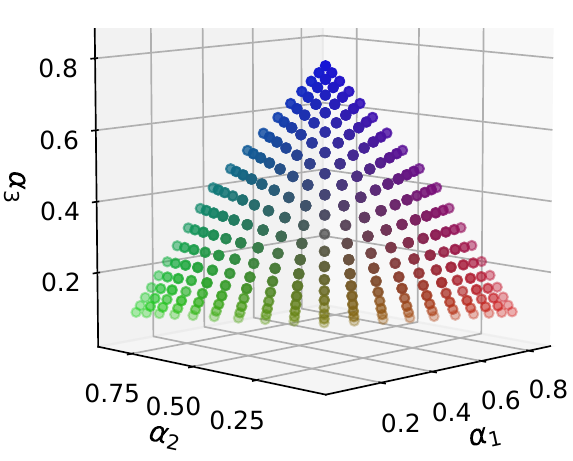}
    \caption{Plotted distribution of weights $\alpha_m$ with $0\leq \alpha_m \leq 1$, $\sum\alpha_m = 1$ and $m=1,2,3$}
    \label{fig:mappedweights}
\end{figure}

By adapting the weights $\alpha_m$ of Equation \ref{eq:gx}, a variety of optimal paths can be achieved, which presents the range of paths that will be interesting to explore. We achieve an equal distribution of weights by logarithmically spacing each weight between 0 and 1 and then normalizing each combination. Figure \ref{fig:mappedweights} shows one example distribution for spacing with step size 10, resulting in $10^3$ combinations. The paths calculated with these combinations create a database which shall be further analysed.

The cost distribution of this database can be mapped in the three-dimensional cost space as shown in one example in Figure \ref{fig:clusteres_imp} with the corresponding path distribution in Figure \ref{fig:clusterspath}.
Not all weight distributions will lead to different paths. Instead, they can be partitioned based on the vicinity of the respective cost in the cost space. 
We use the \emph{greedy k-means++} algorithm, introduced by Arthur and Vassilvitskii \cite{arthur2007k}, to divide the dataset into $k$ groups, where $k$ is user-defined.
To estimate whether the right number of groups ($k$) is found, the user can either visually analyse the plot or examine the variance of each cluster.

\section{RESULTS}
\subsection{Map Data}
\begin{figure*}
    \centering
    \begin{subfigure}{\textwidth}
        \centering
        \includegraphics[width=\linewidth, trim={0 0 0 16cm},clip]{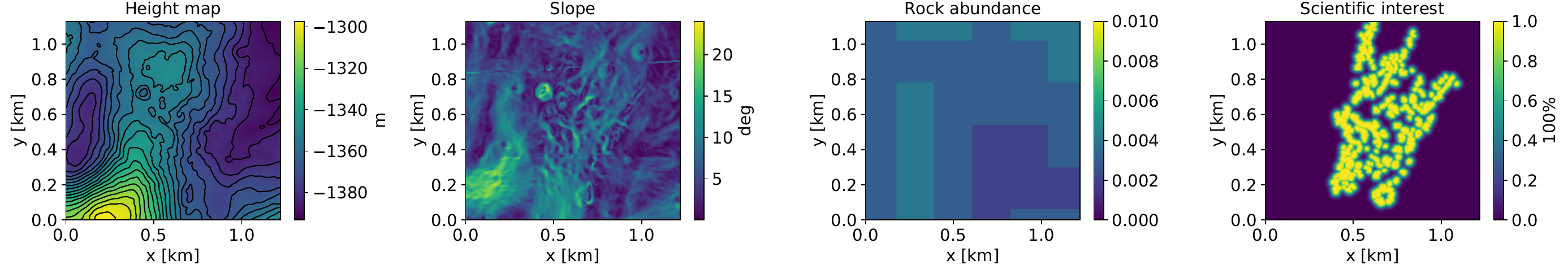}
        \caption{Aristarchus IMP: 256x237 pixel, \unit[4.76]{m/pixel}}
        \label{fig:mapsimp}
    \end{subfigure}
    \begin{subfigure}{\textwidth}
        \centering
        \includegraphics[width=\linewidth, trim={0 16.5cm 0 0},clip]{fig/all_applications_all_maps.pdf}
        \caption{Herodotus Mons: 256x191 pixel, \unit[53.6]{m/pixel}}
        \label{fig:mapshm}
    \end{subfigure}
    \caption{Map layers for two mission scenarios}
    \label{fig:all_maps}
\end{figure*}

For the previously introduced applications, namely the exploration of Herodotus Mons and the Aristarchus IMP, Figure \ref{fig:all_maps} shows the used map layers and Figure \ref{fig:all_applications_sat_pics} shows satellite images of the regions.
The digital elevation model, slope, rock abundance as well satellite image are taken from the \emph{Lunar QuickMap} \cite{lrocquickmap}.

For Herodotus Mons, the scientific interest is based on the concentration of Clinopyroxene, Plagioclase, Iron oxide, and Titanium dioxide, which is data also accessible on the \emph{Lunar QuickMap}. Each value is normalized within its minimum and maximum and summed for each pixel. This map is then smoothed, normalized and thresholded, in order to emphasize areas with elevated element concentrations.

For the Aristarchus IMP, the scientific interests, which include craters, pits, and boulders, are marked manually. This map is then blurred to draw the search algorithm close to the scientific goals. 

\subsection{Run Time Analysis}
The computational cost of the algorithm was measured on an \emph{Intel Core i7-10700KF CPU} running on 8 cores and 16 threads. The planner, which runs in Python, uses $\unit[6.1]{\%}$ of the CPU capacity, independently from the map size. 
A run-time analysis was conducted on the map for Aristarchus IMP (Figure \ref{fig:mapsimp}) and includes scaled map sizes between $64x64$ and $2048x2048$ pixel, five random combinations of start-goal positions as well as the different weight distributions also used in Figure \ref{fig:imppaths}.
The dependence between the run time and the pixel-size can be estimated with:
\begin{equation}
    t = \unit[0.04126]{ms} \cdot n_{pixel} - \unit[69.38]{ms}.
    \label{eq:runtime}
\end{equation}
For the used grid size of \unit[256x256]{pixel}, this leads to an estimated average runtime of \unit[2.634]{s}. Of this, approximately \unit[1.844]{s} is needed to create the maps and initialize the cost and heuristic function and \unit[0.788]{s} to run the planner.

The run time is further dependent on the number of map layers and the complexity of the cost function. For example for the proposed cost, each step involves the calculation of Equations \ref{eq:wells1}, \ref{eq:wells3} and \ref{eq:r_scaled}. Consequently, how each cost is computed has a stronger impact on the runtime than the actual number of map layers, making it challenging to give a general estimate. 

\subsection{General Performance Analysis}
\begin{figure*}
    \centering
    \begin{subfigure}{.32\textwidth}
        \centering
        \includegraphics[width=\linewidth,trim={1.25cm 0 0 0},clip]{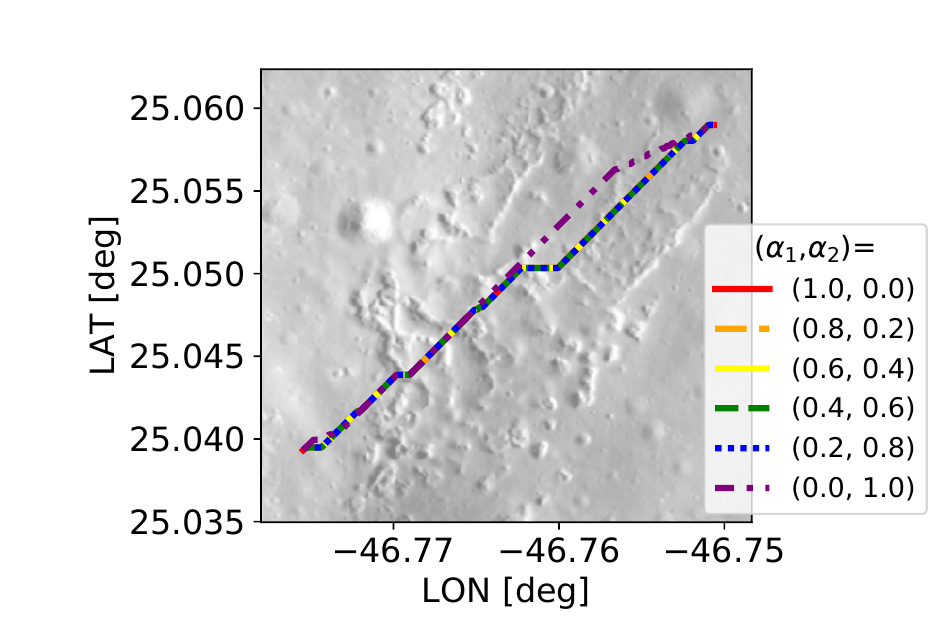}
        \caption{Energy ($\alpha_1$) vs. Risk ($\alpha_2$)}
        \label{fig:vs1}
    \end{subfigure}
    \hfill
    \begin{subfigure}{.32\textwidth}
        \centering
        \includegraphics[width=\linewidth,trim={0 0 1.2cm 0},clip]{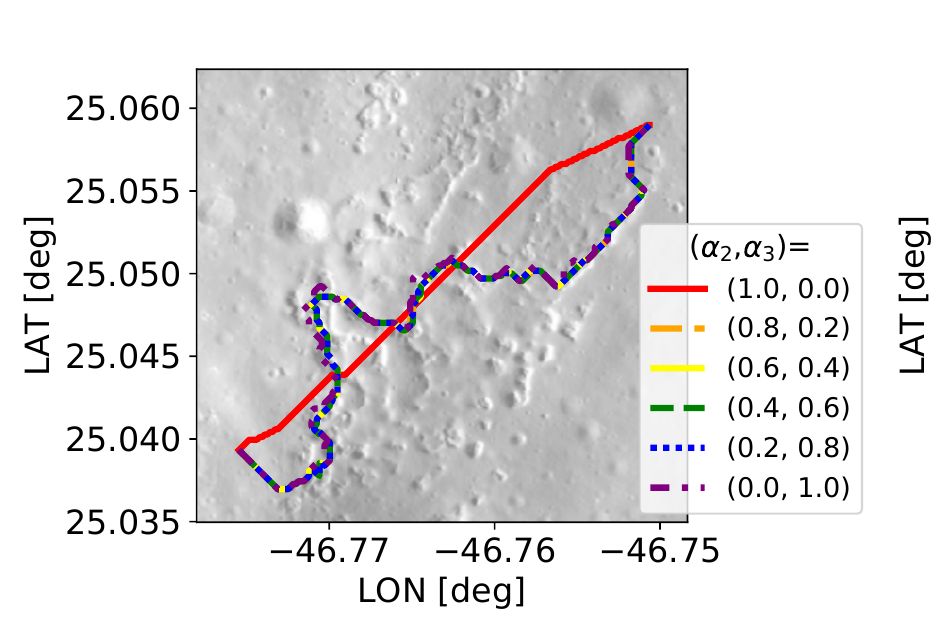}
        \caption{Risk ($\alpha_2$) vs. Science ($\alpha_3$)}
        \label{fig:vs2}
    \end{subfigure}
    \hfill
    \begin{subfigure}{.32\textwidth}
        \centering
        \includegraphics[width=\linewidth]{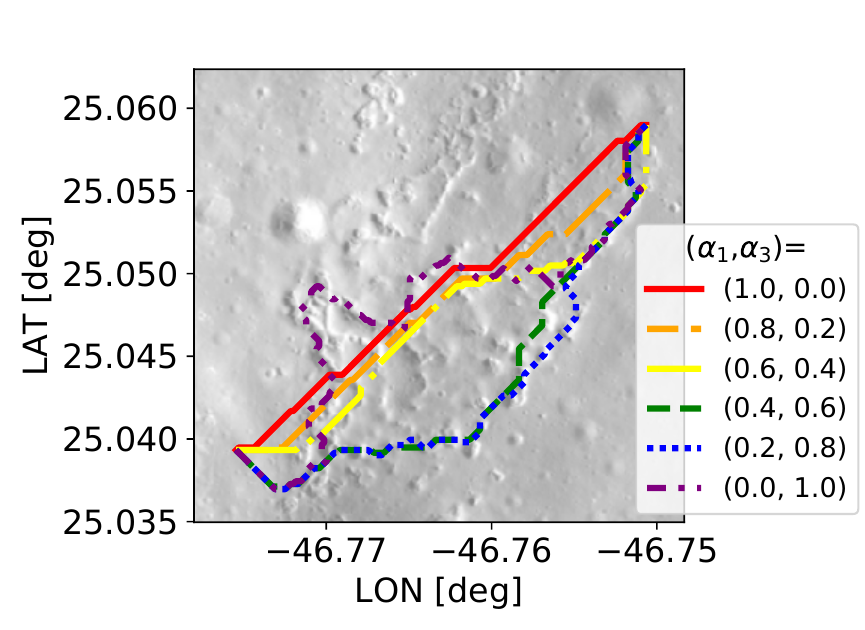}
        \caption{Energy ($\alpha_1$) vs. Science ($\alpha_3$)}
        \label{fig:vs3}
    \end{subfigure}
    \caption{Results of path planning for selected weights on Aristarchus IMP}
    \label{fig:imppaths}
\end{figure*}

Figure \ref{fig:imppaths} displays the resulting paths for the Aristarchus IMP, visualized on a satellite image and based on the previously introduced maps (Figure \ref{fig:mapsimp}). Each subfigure depicts trajectories that result from incremental variations in two of the three weights, respectively. Table \ref{tab:costs256} calculates the cost when considering only one of the three objectives at a time.

Figure \ref{fig:vs1} presents the synergy between energy efficiency ($\alpha_1$) and risk ($\alpha_2$). It is evident that the paths are very similar with minor variations towards the end. Upon revisiting the modelled risk and energy efficiency it becomes apparent that this similarity is expected. Both cost objectives are dependent on the slope and the rock abundance with minima close to $(s,r) = (0,0)$, thereby guiding the path near these values.

Figure \ref{fig:vs2} presents the range of path results for varied focus on risk or scientific significance. The six different settings manifest themselves in two distinct paths. 
This observation suggests that the cost of taking risk does not have a significant influence on the overall cost, a fact that is confirmed by the data in Table \ref{tab:costs256}. Specifically, the cost values associated with risk are below $1$ for the three corner cases, whereas the costs related to energy efficiency and scientific value are significantly higher.
The explanation can again be found in the employed crash rate. Given the mostly flat terrain with a low rock abundance, the cost for the risk remains in proximity to its minima 0.

\begin{table}[bp]
    \centering
    \caption{Selected costs for different weights on Aristarchus IMP}
    \label{tab:costs256}
    \begin{tabular}{ll|rrr}
          & & \multicolumn{3}{l}{($\alpha_1$, $\alpha_2$, $\alpha_3$) =} \\
         \multicolumn{2}{l|}{Cost Components} & (1,0,0) & (0,1,0) & (0,0,1) \\
         \hline \hline
         Energy & $E$ & 65.3 & 65.6 & 115.1 \\
         Risk & $R$ &0.00540 & 0.00539 & 0.4557 \\
         Scientific & $I$ & 82.2 & 100.9 & 38.0 \\
         \hline
         Total & $\alpha_1 E + \alpha_2 R + \alpha_3 I$ & 65.3 & 0.00539 & 38.0 \\
         \hline \hline
         Path length & [km] & 0.758 & 0.758 & 1.38 
    \end{tabular}
    \vspace{0.5cm}
    \caption{Selected costs for different weights on Herodotus Mons}
    \label{tab:hmcosts}
    \begin{tabular}{ll|rrr}
         & & \multicolumn{3}{l}{($\alpha_1$, $\alpha_2$, $\alpha_3$) =} \\
         \multicolumn{2}{l|}{Cost Components} & (1,0,0) & (0,1,0) & (0,0,1) \\
         \hline \hline
         Energy & $E$ & 53.5 & 60.7 & 55.9 \\
         Risk & $R$ & 23.0 & 0.0133 & 47.0 \\
         Scientific & $I$ & 100.5 & 141.6 & 73.2 \\
         \hline
         Total & $\alpha_1 E + \alpha_2 R + \alpha_3 I$ & 53.5 & 0.0133 & 73.2 \\
         \hline \hline
         Path length & [km] & 6.92 & 8.58 & 5.31 
    \end{tabular}
\end{table}

\begin{figure*}
    \centering
    \begin{subfigure}{.32\textwidth}
        \centering
        \includegraphics[width=.94\linewidth, trim={0 0 0 0.8cm}, clip]{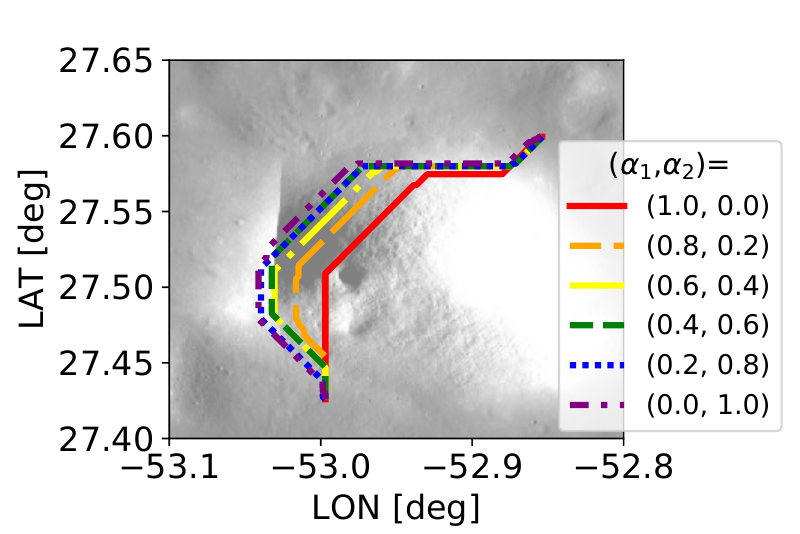}
        \caption{Energy ($\alpha_1$) vs. Risk ($\alpha_2$)}
        \label{fig:vs21}
    \end{subfigure}
    \hfill
    \begin{subfigure}{.32\textwidth}
        \centering
        \includegraphics[width=.94\linewidth, trim={0 0 0 2.3cm}, clip]{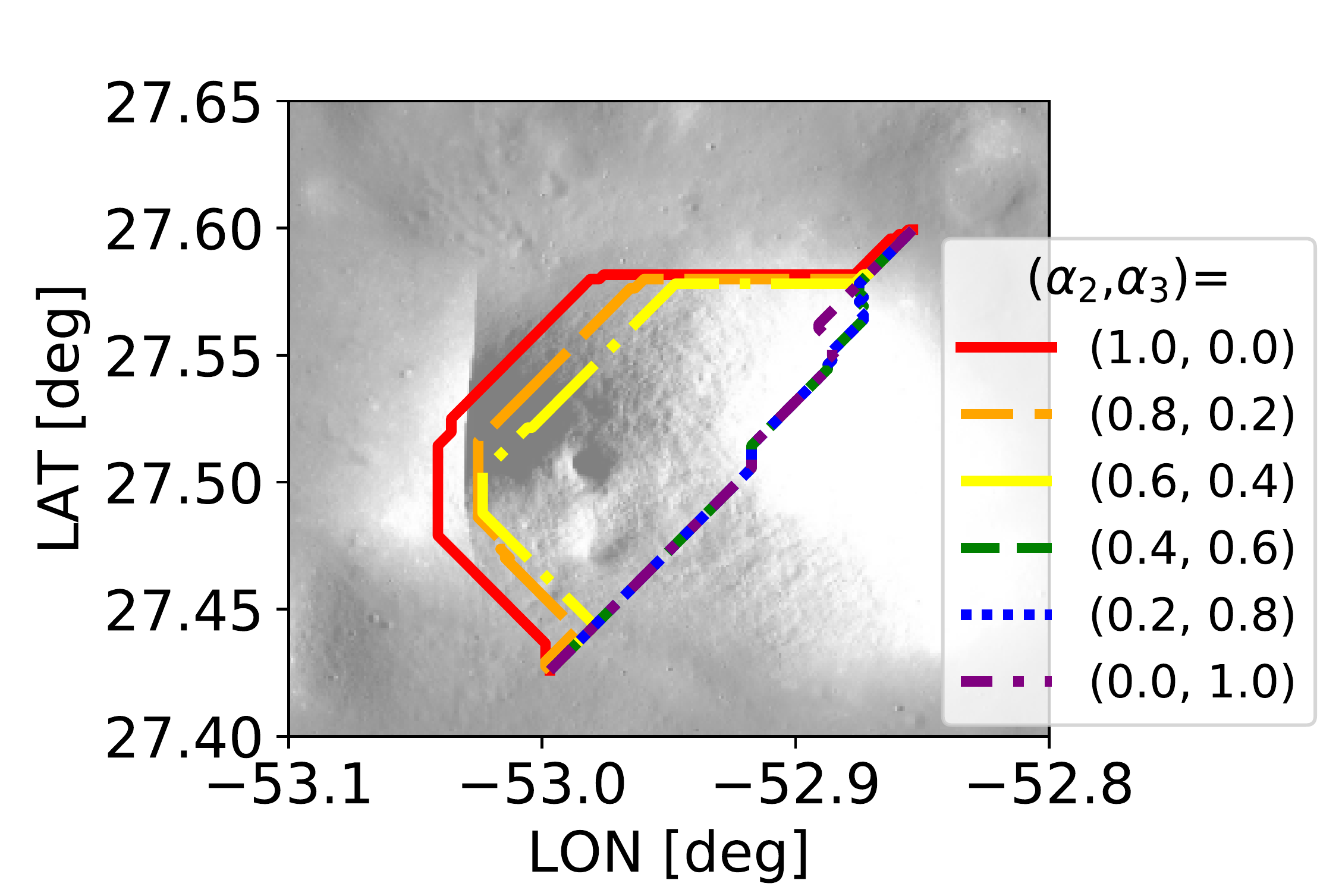}
        \caption{Risk ($\alpha_2$) vs. Science ($\alpha_3$)}
        \label{fig:vs22}
    \end{subfigure}
    \hfill
    \begin{subfigure}{.32\textwidth}
        \centering
        \includegraphics[width=.94\linewidth, trim={0 0 0 2.3cm}, clip]{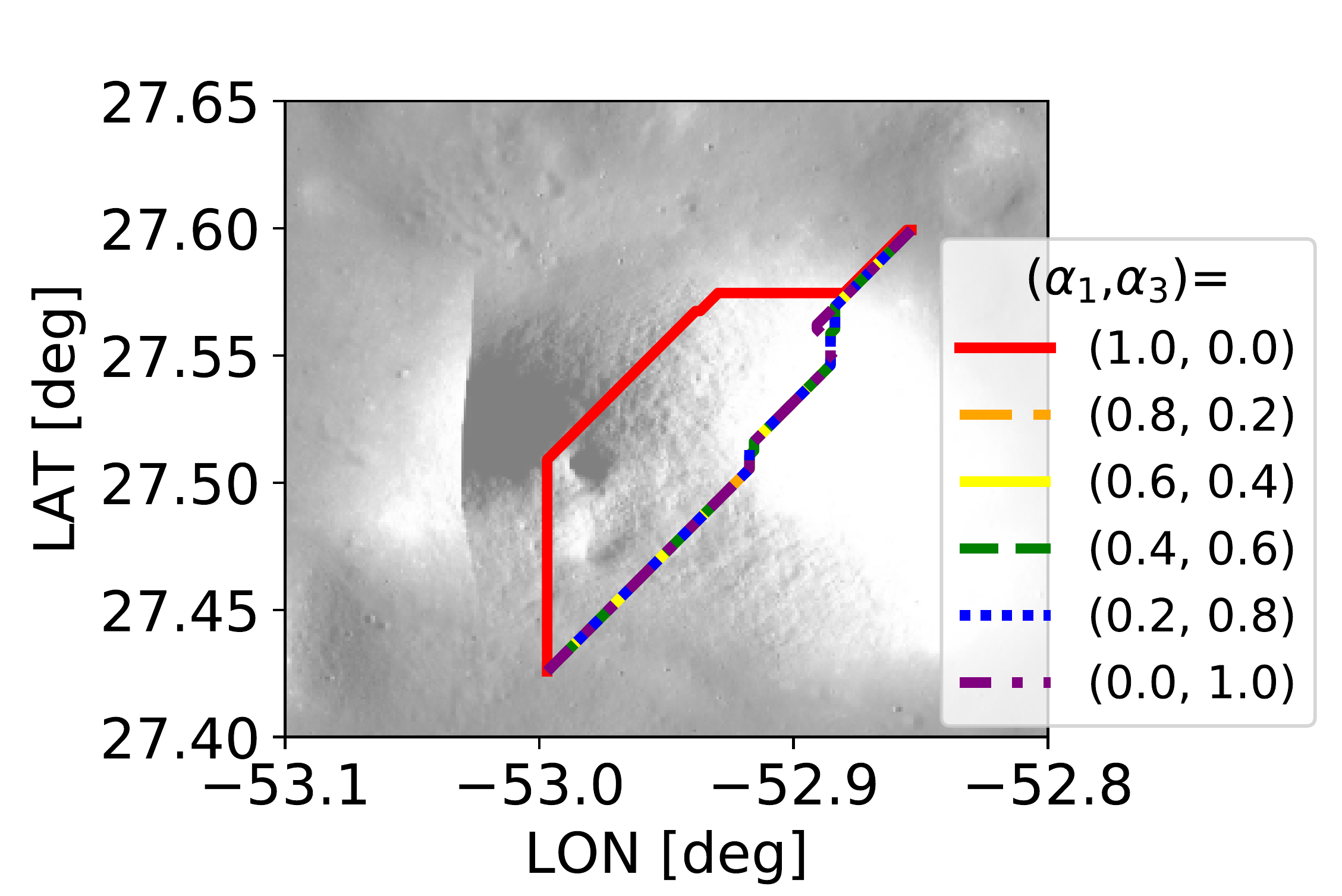}
        \caption{Energy ($\alpha_1$) vs. Science ($\alpha_3$)}
        \label{fig:vs23}
    \end{subfigure}
    \caption{Results of path planning for selected weights on Herodotus Mons}
    \label{fig:hmpaths}
\end{figure*}

While the first subplot in Figure \ref{fig:vs1} showed similar paths, and Figure \ref{fig:vs2} exhibits a bi-stable result, \ref{fig:vs3} demonstrates a gradual transition.
The path evolves from a focus on maximizing scientific value to prioritizing energy efficiency, gradually reducing detours and lowering the relative cost for energy efficiency from $115.1$ to $65.3$, as detailed in Table \ref{tab:costs256}.

Figure \ref{fig:hmpaths} and Table \ref{tab:hmcosts} present the same path details for an example exploration of Herodotus Mons. 
While the previous example showed a strong dependency on the chosen cost function, this example highlights the impact of diverse map appearances.
By revisiting Figures \ref{fig:mapsimp} and \ref{fig:mapshm}, we can see that the rock abundance and scientific interest for the two applications look very different. While the Aristarchus IMP has distinctive scientific targets and, therefore, many pixels with a value $1$, for Herodotus Mons this layer has mainly values between $0.4$ and $0.7$. The rock abundance on the contrary has a low resolution and magnitude for Aristarchus IMP and a structured appearance for Herodotus Mons.
Lastly, the height map shows a generally higher slope for Herodotus Mons than for Aristarchus IMP.

The higher magnitudes for rock abundance and slope of Herodotus Mons lead to a generally higher crash risk $R^*$ and hence higher values for the cost component $R$, as shown in Table \ref{tab:hmcosts}. Consequently, Figure \ref{fig:vs21} and \ref{fig:vs22} show more distinctions than Figure \ref{fig:vs1} and \ref{fig:vs2}.
At the same time, the scientific cost per traversed meter is lower, which leads to less influence of the weight $\alpha_3$ and hence a bi-stable appearance of the paths in Figure \ref{fig:vs23}.

\begin{table}[b]
    \centering
    \caption{Aristarchus IMP: Details of calculated clusters}
    \label{tab:details_clusters_imp}
    \begin{tabular}{r|c c c| c c}
         Path & Energy & Risk & Science & No. paths & Cluster \\
         No. & [\%] & [\%] & [\%] & in cluster & variance \\
         \hline
         1 & 78.2 & 0.95 & 57.02 & 58 & 0.03 \\ 
         2 & 81.1 & 3.73 & 67.42 & 262 & 0.73 \\ 
         3 & 87.7 & 0.23 & 76.38 & 201 & 0.7 \\ 
         4 & 100 & 17.82 & 81.94 & 326 & 1.84 \\ 
    \end{tabular}
    \vspace{0.5cm}
    \caption{Herodotus Mons: Details of calculated clusters}
    \label{tab:details_clusters_hm}
    \begin{tabular}{r|c c c| c c}
         Path & Energy & Risk & Science & No. paths & Cluster \\
         No. & [\%] & [\%] & [\%] & in cluster & variance \\
         \hline
         1 & 100 & 26.63 & 21.64 & 91 & 1.94 \\ 
         2 & 98.6 & 52.08 & 24.63 & 343 & 0.54 \\ 
         3 & 92.7 & 98.1 & 24.04 & 28 & 0.22 \\ 
         4 & 94.7 & 100.0 & 26.05 & 385 & 0.04 \\ 
    \end{tabular}
\end{table}

\subsection{Statistical Path Analysis}
\begin{figure}
    \centering
    \begin{subfigure}[b]{.47\linewidth}
        \centering
        \includegraphics[width=\linewidth]{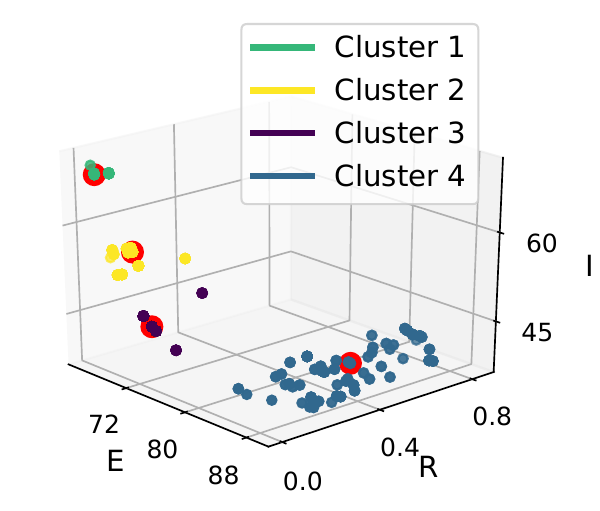}
        \caption{Aristarchus IMP: Cost values}
        \label{fig:clusteres_imp}
    \end{subfigure}
    \hfill
    \begin{subfigure}[b]{.51\linewidth}
        \centering
        \includegraphics[width=\linewidth]{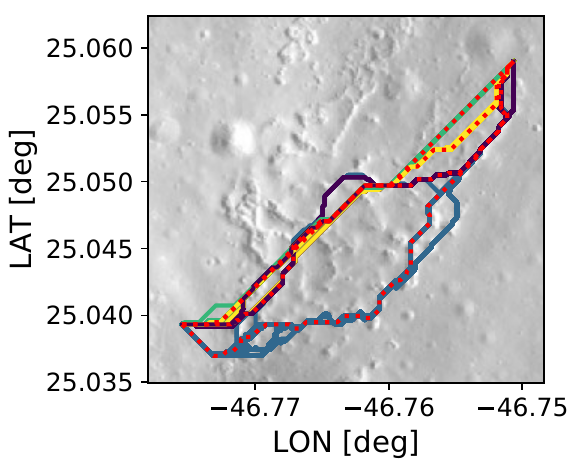}
        \caption{Aristarchus IMP: Resulting paths}
        \label{fig:clusterspath}
    \end{subfigure}

    \begin{subfigure}[b]{.44\linewidth}
        \centering
        \includegraphics[width=\linewidth]{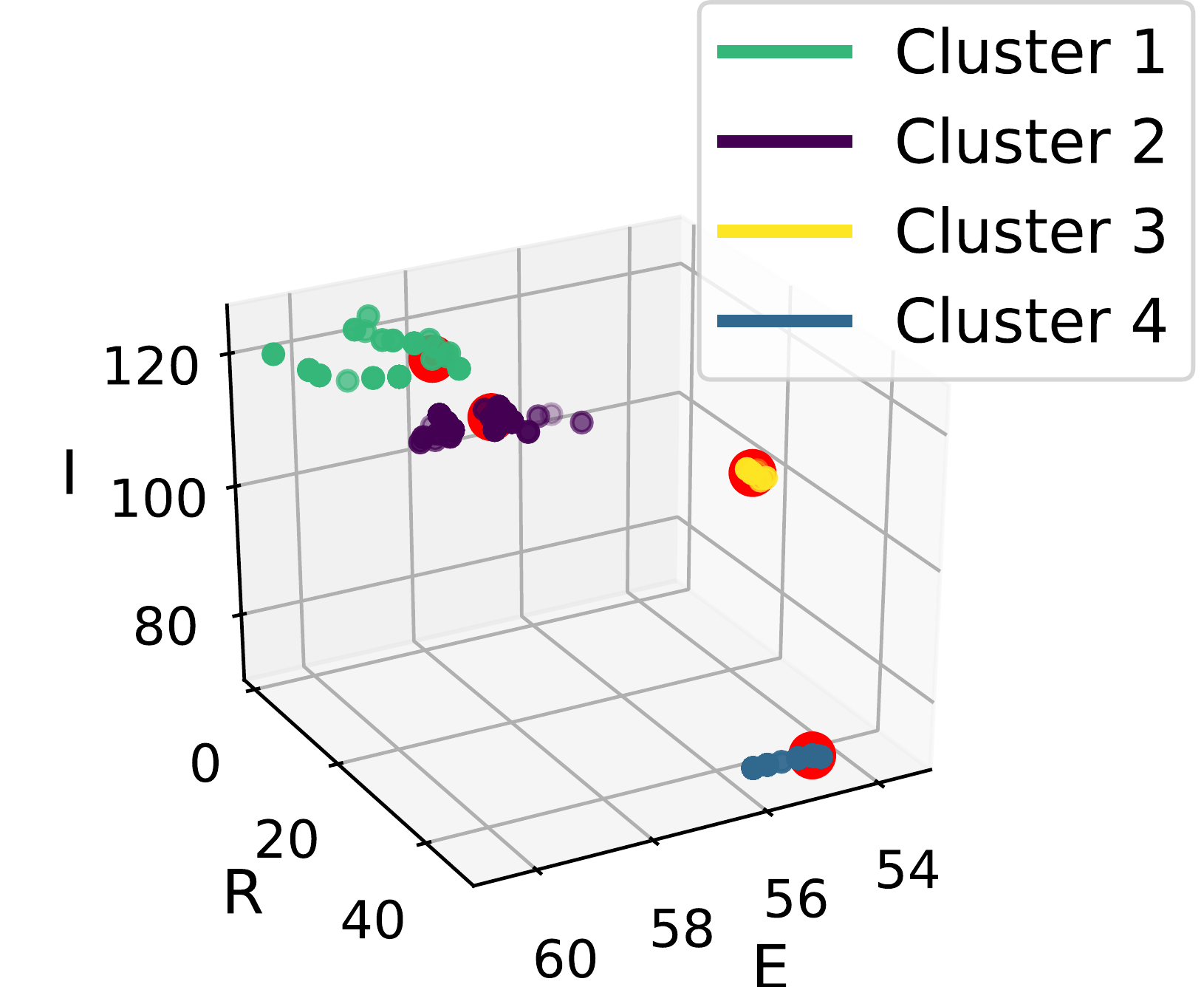}
        \caption{Herodotus Mons: Cost values}
        \label{fig:clusteres_hm}
    \end{subfigure}
    \hfill
    \begin{subfigure}[b]{.54\linewidth}
        \centering
        \includegraphics[width=\linewidth]{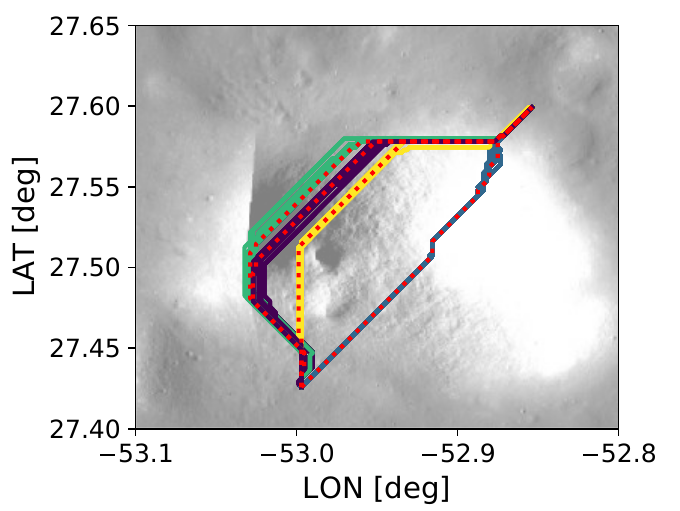}
        \caption{Herodotus Mons: Resulting paths}
        \label{fig:clusterspath_hm}
    \end{subfigure}
    \caption{Resulting path clusters in different views. The center value is highlighted in red.}
    \label{fig:clusteranalysis}
\end{figure}

The previously selected weight distributions only represent an extract of possible paths. A statistical analysis as described in Section \ref{sec:methods_analysis} can be done to understand the influence of the map on the result and find the optimal path for each application. Initially, a database with paths resulting from all 1000 weight combinations needs to be created. This took \unit[15.4]{min} for the paths on Herodotus Mons (Figure \ref{fig:clusterspath}) and \unit[12.6]{min} for Aristarchus IMP (Figure \ref{fig:clusterspath_hm}).

Figures \ref{fig:clusteres_imp} and \ref{fig:clusteres_hm} show the three-dimensional cost space. The different colours represent the chosen clusters. The number of clusters $k=4$ was optimized based on the cost-variance of each cluster as well as the visible path variance in Figures \ref{fig:clusterspath} and \ref{fig:clusterspath_hm}. The red highlight is the distribution closest to the respective center of each cluster. Figures \ref{fig:clusterspath} and \ref{fig:clusterspath_hm} show all calculated paths in the same coloring as the cost plot. The path belonging to the most centered cost is again highlighted in red. 
Table \ref{tab:details_clusters_imp} and \ref{tab:details_clusters_hm} present the resulting physical specs as well as details about the cluster quality. They are calculated using Equations \ref{eq:etotal}, \ref{eq:rtotal}, and \ref{eq:itotal}. 
Since the energy efficiency $E^*_{total}$ is only a relative estimate, as explained earlier, it is scaled to the respective maximum value of each mission.

The physical properties of energy efficiency, risk of failure, and scientific outcome are directly related to the respective cost values. 
A low cost for E leads to a lower relative energy. 
This can be seen in cluster 1 in of Aristarchus IMP (Figure \ref{fig:clusteres_imp}). A lower cost for R, like for cluster 1 of Herodotus Mons (Figure \ref{fig:clusteres_hm}), leads to a lower risk. And lastly, a lower cost for I, like for cluster 4 of Aristarchus IMP (Figure \ref{fig:clusteres_imp}), leads to a higher scientific outcome.

In order to choose an optimal path for Aristarchus IMP, the energy and scientific outcome have a stronger influence than the risk. Clusters 3 and 4 have only slightly different values for Energy ($\unit[87.7]{\%}$ and $\unit[100]{\%}$) and Science ($\unit[76.38]{\%}$ and $\unit[81.94]{\%}$) while resulting in completely different paths. One mission strategy could be to follow path 3 and consecutively take path 4 back to the initial position since the higher risk of failure on path 4 ($\unit[17.82]{\%}$ vs. $\unit[0.23]{\%}$) can be taken towards the end of a mission.

In contrast to that, for Herodotus Mons, the risk of failure has a very high impact. Only paths 1 and 2 can be considered since paths 3 and 4 have an extremely high risk of failure ($\unit[98.1]{\%}$ and $\unit[100.0]{\%}$). Between the remaining options, path 1 only has a minor scientific loss ($\unit[2.99]{\%}$) and slightly more relative energy consumption ($\unit[1.4]{\%}$), while cutting the risk in half. So choosing path 1 for exploration is recommended.

\subsection{Proposed Exploration Missions} \label{sec:proposed}
In the following, two specific missions to explore the Herodotus Mons region and the Aristarchus IMP are proposed.
For both scenarios, a lunar geology expert previously planned traverses for a legged robot. The here proposed method inherits the lander and final position as well as the approximate course of the path. 
Additionally, the following constraints should be considered:
\begin{itemize}
    \item The relative energy consumption must not exceed the manually planned path.
    \item The risk of failure should be minimized.
    \item The scientific outcome should be maximized.
\end{itemize}

The approximate courses of the paths were generated with waypoints, which are highlighted with red dots in Figures \ref{fig:imp_proposed} and \ref{fig:hm_proposed}.
The generation of the path database took \unit[10.0]{min} for Herodotus Mons and \unit[36.1]{min} for Aristarchus IMP.
Using the above-stated constraints, the analysis tool was then used to find the best weight distribution for different path segments. The resulting paths are shown in Figure \ref{fig:all_applications_sat_pics} together with the manually planned paths. Table \ref{tab:comparisonpaths} presents the details of each result.
The relative estimate for the energy efficiency $E^*_{total}$ is scaled to the value of the manually planned path of each mission.

Since the trajectory of the tool-assisted planning was inspired by the manual path, the energy consumption is very similar. However, for Herodotus Mons, it exceeds the manually planned path by $5.2\%$. This is attributed to serpentines proposed by the planner in the steep section highlighted in blue in Figure \ref{fig:hm_proposed}. While opting for a more direct ascent could save energy, it would significantly elevate the risk. 
For Aristarchus IMP, the tool-assisted planning yields a significantly higher scientific outcome (\unit[+36.3]{\%}) with a reduced risk (\unit[-45.6]{\%}).

Although the human planner could further optimize the manually planned path to improve output values, doing so would require substantially more time and might not consistently achieve the optimal balance between the objectives.

\begin{table}
    \centering
    \caption{Results for paths planned manually and tool-assisted}
    \begin{tabular}{ll|rr|r}
        \multicolumn{5}{c}{Aristarchus IMP} \\
        & & Manual & Tool-assisted & \\
        \hline
        Length & [$km$] & 3.23 & 3.22 & $-0.3\%$ \\
        Energy & [$\%$] & 100 & 99.94 & $-0.06\%$ \\
        Risk & [$\%$] & 36.0 & 19.6 & $-45.6\%$ \\
        Science & [$\%$] & 37.6 & 51.3 & $+36.3\%$ \\
        \hline
        \multicolumn{5}{c}{Herodotus Mons} \\
        & & Manual & Tool-assisted & \\
        \hline
        Length & [$km$] & 8.58 & 9.26 & $+8.3\%$ \\
        Energy & [$\%$] & 100 & 105.2 & $+5.2\%$ \\
        Risk & [$\%$] & 95.3 & 50.4 & $-47.1\%$ \\
        Science & [$\%$] & 27.3 & 48.7 & $+78.4\%$
    \end{tabular}
    \label{tab:comparisonpaths}
\end{table}

\section{DISCUSSION}
\subsection{Limits of the Global Planner:  Runtime and Accuracy}
The runtime of the planner is linearly dependent on the number of pixels. For larger map sizes, the runtime expands, which leads to an inefficient planning process. For such cases, it is suggested to specify an approximate path, as demonstrated in Section \ref{sec:proposed}.

The accuracy of the result from the global planner depends on several factors. The most important are accurate models for the cost. While the here used costs were estimated in simulation, exemplary tests in reduced gravity should be explored to close the sim-to-real gap. 

Additionally, the here implemented cost objectives are just one example of what is possible and the planner can be easily extended. Since the cost is calculated at every step, dynamically changing maps like illumination or telecommunication quality can potentially be integrated in the future. However, the complexity of the cost should be taken into consideration to limit the run time.

\subsection{Limits of the Analysis Tool:  Runtime}
To establish a database for the analysis tool, the planner must be run with numerous weight distributions. For longer missions, this process took up to \unit[36]{min}. 
This duration is acceptable if the database creation is executed once, meaning that the intermediate waypoints remain fixed.
However, it is impractical if various start and goal positions need to be considered. One apparent solution would be to create a smaller database than the here selected 1000 weight combinations. Looking at the density of resulting clusters, we assume that this will not have a negative impact on the path quality for most applications and should hence be further investigated.

\section{CONCLUSION}
This work developed a global path planner by adapting the well-known A* algorithm. 
The new planner is capable of separately considering multiple layers of map data for a multi-objective cost function. Each objective is assigned a weight, which allows operators to adapt the calculated paths while ensuring optimality. Additionally, a tool was developed to provide statistical analysis of the resulting paths. This analysis aids in selecting the most suitable path for a given application.
The quality of the resulting paths highlight the potential of automating global planning.

This work took integral steps towards simplification and acceleration in proposing trajectories within the mission planning phase.
Our code will be made available as an easy-to-use tool for planetary scientists to plan future missions of planetary exploration.
Currently, only the quadruped robot ANYmal with its respective cost functions is implemented. However, other robots and rovers can be easily added.

\section{FUTURE WORK}
So far, the global planner was only tested in simulation for lunar applications. In order to validate the proposed method, field trials on Earth in lunar-like environments should be done. This necessitates a full navigation pipeline, including global localisation as well as local perception and planning.
Additionally, we will be integrating local perception information to update global paths, for instance, when the locally detected rock abundance exceeds the previous estimate.

Lastly, future planetary exploration will probably be conducted with teams of heterogenuous robots. The next step for this planner is to optimise the usage for several robots with different locomotion or science capacities.

\addtolength{\textheight}{-12cm}   




\section*{ACKNOWLEDGMENT}
We acknowledge the use of imagery from Lunar QuickMap (https://quickmap.lroc.asu.edu), a collaboration between NASA, Arizona State University \& Applied Coherent Technology Corp.


\bibliographystyle{IEEEtran}
\bibliography{bibtex}

\end{document}